# SocioSense: Robot Navigation Amongst Pedestrians with Social and Psychological Constraints

Aniket Bera and Tanmay Randhavane and Rohan Prinja and Dinesh Manocha[1]

*Abstract*— We present a real-time algorithm, *SocioSense*, for socially-aware navigation of a robot amongst pedestrians. Our approach computes time-varying behaviors of each pedestrian using Bayesian learning and Personality Trait theory. These psychological characteristics are used for long-term path prediction and generating proxemic characteristics for each pedestrian. We combine these psychological constraints with social constraints to perform human-aware robot navigation in low- to medium-density crowds. The estimation of time-varying behaviors and pedestrian personalities can improve the performance of long-term path prediction by 21%, as compared to prior interactive path prediction algorithms. We also demonstrate the benefits of our socially-aware navigation in simulated environments with tens of pedestrians.

## I. INTRODUCTION

Robots are increasingly used in households, offices and public places and frequently navigate amongst humans or pedestrians. As humans are dynamic agents, these scenarios result in many new challenges related to *human-aware navigation and interaction*. Robots must move through crowds of people while preventing collisions with each other and with humans. In such scenarios, the robots need to interface with not only the physical environment, but also the social environment, and should interact well with humans.

People have clear social norms about interpersonal space or acceptable behaviors [6]. A robot that impinges on someone's personal space could make them feel uncomfortable [34]. There is considerable work on socially-aware navigation in motion planning and HRI (human-robot interaction) literature. These techniques tend to predict the movement or actions of human pedestrians and use them to develop appropriate navigation algorithms [19], [25], [11], [20]. The resulting paths account for social norms and conventions, such as personal space and yielding the right of way.

In this paper, we address the problem of a robot navigating through low- and medium-density crowds. In addition to social constraints, we also take into account the personality or time-varying behaviors of different pedestrians. It is known in psychology that even a simple task such as walking towards a destination involves several complex human-centric decisions, such as which route to take and the various ways to avoid collision with the robot, other pedestrians, and the obstacles in the environment. In many scenarios, different pedestrians will accomplish the same goal in different ways and their resulting paths are governed by their underlying personality (e.g. aggressive, shy, or active). Therefore, it is important to capture the pedestrians' personality traits, and use them to predict their future positions for socially-aware navigation. In many dense crowd scenarios, the speed and movement pattern of a pedestrian may change in response to environmental factors, such as crowd density and overall flow. A key issue is therefore modeling variations in behavior that arise as humans navigate in the physical world and avoid collisions. Furthermore, it is important to capture and compute these time-varying behaviors at interactive rates, so that the robot can navigate accordingly.

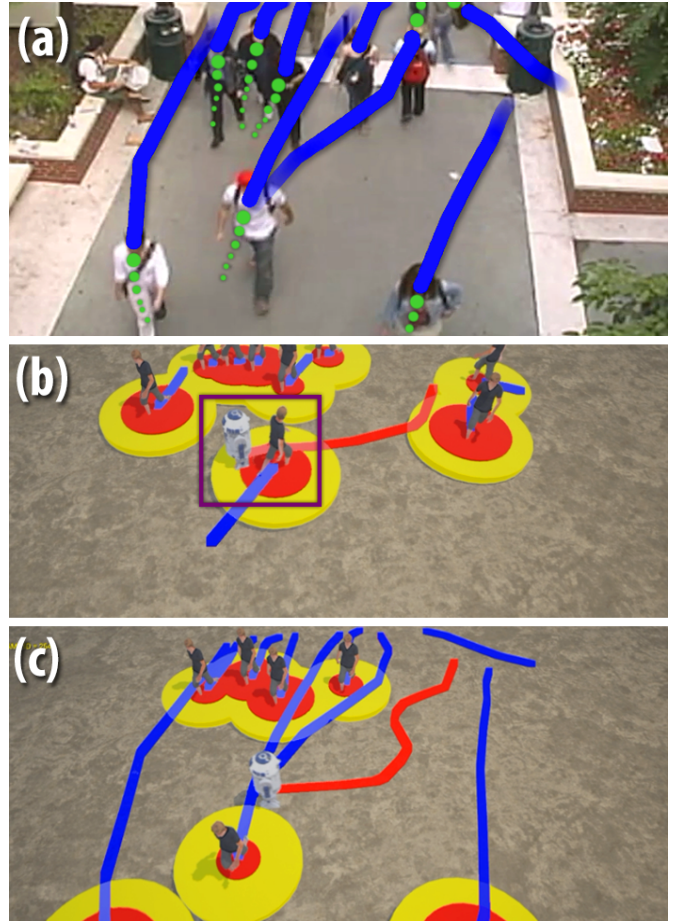

**Fig. 1:** *Improved Navigation using **SocioSense**: (a) shows a real-world crowd video and the extracted pedestrian trajectories in blue. The green markers are the predicted positions of each pedestrian computed by our algorithm that are used for collision-free navigation. The red and yellow circles around each pedestrian in (b) and (c) highlight their personal and social spaces respectively, computed using their personality traits. We highlight the benefits of our navigation algorithm that accounts for psychological and social constraints in (c) vs. an algorithm that does not account for those constraints in (b). The red trajectory of the white robot maintains these interpersonal spaces in (c), while the robot navigates close to the pedestrians in (b) and violates the social norms.*

[1]All the authors from the Department of Computer Science, University of North Carolina at Chapel Hill, USA

**Main Results:** We present a real-time planning algorithm, *SocioSense*, that takes into account psychological constraints of each pedestrian in the environment to perform socially-aware navigation. We extract the trajectory of each pedestrian from the video and use Bayesian learning algorithms to compute his/her motion model and the personality characteristics of the pedestrian. Our behavior classification is based on Personality Trait Theory in psychology literature. This theory assumes that the variations in behavior are governed by a small number of underlying traits.

We combine the time-varying behavior classifications for each pedestrian with local and global learning methods to perform long-term path prediction for collision-free, socially normative robot navigation. A key contribution of this paper is mapping the learned personalities to a set of social distances based on proxemics and interpersonal distances [15]. These distances constrain the robot navigation to avoid passing through people's personal and social spaces. Using the combination of psychological and social constraints improves both the prediction and the navigation. We have evaluated the performance of our algorithm in real-world captured videos consisting of tens of pedestrians, including dense scenarios. Some of the main benefits of our approach include:

As compared to prior pedestrian learning and social navigation algorithms, *SocioSense* offers the following benefits:

**1. Robust:** *SocioSense* is general and can account for noise in the extracted pedestrian trajectories. It also works in scenarios corresponding to low- and medium- density crowds and can capture the personality of each pedestrian.

**2. Fast and Accurate:** *SocioSense* involves no precomputation and evaluates the time-varying behaviors at interactive rates. The use of psychological constraints improves the accuracy of long-term pedestrian prediction by 21% over prior interactive methods.

**3. Socially-Aware Robot Navigation:** In addition to ensuring safe navigation of robots around pedestrians, *SocioSense* also takes into account their personalities and social distances to generate motion that is comfortable to the pedestrians.

## II. RELATED WORK

In this section, we give a brief overview of prior work on robot navigation in human environments and socially-aware navigation.

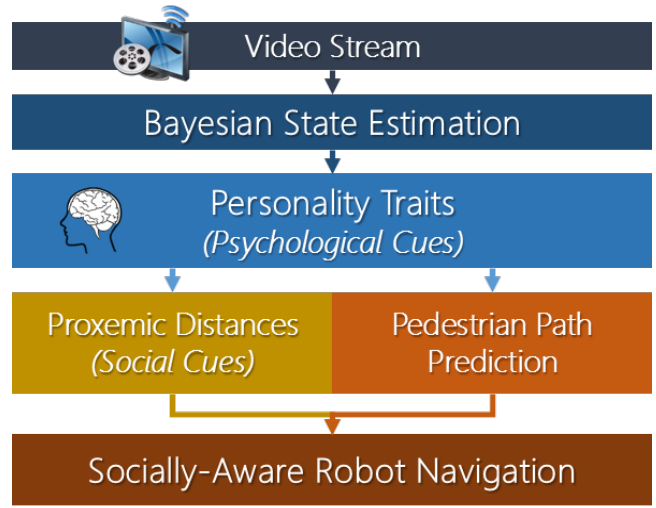

**Fig. 2: Overview of *SocioSense*:** *Our method takes a streaming crowd video as an input. We compute the state of pedestrians in the crowd from extracted trajectories, as explained in Section III. Based on the state information, we learn the pedestrians' personality traits, which are used to improve pedestrian prediction and to compute proxemic/interpersonal distances. The proxemic distances and predicted trajectories are combined with a collision-avoidance algorithm to perform socially-aware robot navigation.*

### A. Robot Navigation Amongst Pedestrians

There is considerable work on robot navigation in human environments. This work includes many early systems such as RHINO [5] and MINERVA [35], which use robots as tour guides in museums, and other systems for robot navigation in urban environments [2], [21]. Some navigation algorithms use potential-based approaches for robot path planning in dynamic environments [29]. Different techniques have been proposed for pedestrian trajectory prediction using probabilistic or Bayesian velocity-obstacles [12], [17] and a partially closed loop receding horizon control [7]. Other methods are based on learning the motion characteristics or similar data from real-world trajectories, including using Markov decision processes to model human behaviors [40] and to predict future pedestrian trajectories [3], as well as using reinforcement learning to imitate human behavior [16]. Our approach uses psychological and social constraints, and can be combined with most of these methods to perform human-aware navigation.

### B. Social Robot Navigation

There is considerable work on social robot navigation and human-aware navigation [26], [19], [25], [11], [20]. Some navigation algorithms generate socially compliant trajectories by predicting the pedestrian movement and forthcoming interactions [20] or use modified interacting Gaussian processes to develop a probabilistic model of robot-human cooperative collision avoidance [36]. Other methods tend to model interactions and personal space for human-aware navigation [1] or use learning-based approaches to account for social conventions in robot navigation [22], [27], [30]. Many explicit models have been proposed for social constraints to enable person-acceptable navigation [33], [18]. However, these methods do not take into account psychological constraints or personalities of the pedestrians.

## III. PSYCHOLOGICAL CONSTRAINTS AND PATH PREDICTION

In this section, we introduce the notation and present our approach to model psychological and social constraints. In human environments, it is important to compute robot trajectories that follow the social norms about interpersonal distance and acceptable behaviors. Most of the earlier work is limited to collision-free trajectory planning or only takes into account physical constraints such as kinematics and dynamics. The psychological constraints are complimentary to these physical constraints and depend on the personality or behavior of a pedestrian in the crowd.

Fig. 2 gives an overview of our approach, including computation of the different constraints and their use in navigation. Our real-time algorithm learns pedestrians' psychological and social cues from trajectories extracted from real-world crowd videos. Our approach is generic and can be combined with almost any real-time pedestrian tracker that works well on low- to medium- density crowds. We learn a pedestrian's personality traits (psychological constraints) from the trajectories using Bayesian learning. We use these time-varying psychological constraints to compute appropriate interpersonal distances for each pedestrian for robot navigation (social constraints). The personality classification is also used for long-term path prediction, which is also used by the socially-aware robot navigation algorithm.

### A. Symbols and Notation

We introduce the terminology and symbols used in the rest of the paper. We use lowercase letters for scalars and bold letters for vectors. We refer to an agent in the crowd as the *pedestrian* whose *state* includes his/her trajectory and behavior characteristics. This state, denoted by the symbol $\mathbf{x} \in \mathbb{R}^5$, governs the pedestrian's position on the 2D plane:

$$\mathbf{x} = [\mathbf{p} \ \mathbf{v}^c \ \mathbf{v}^{pref}]^\mathbf{T}; \quad (1)$$

where $\mathbf{p}$ is the pedestrian's position, $\mathbf{v}^c$ is his/her current velocity, and $\mathbf{v}^{pref}$ is the *preferred velocity* on a 2D plane. A pedestrian's current velocity $\mathbf{v}^c$ tends to be different than the optimal velocity (the preferred velocity $\mathbf{v}^{pref}$) that he/she would take in the absence of other pedestrians or obstacles in the scene to achieve his/her intermediate goal. The union of the states of all the other pedestrians and the current positions of the obstacles in the scene is the current state of the environment denoted by the symbol $\mathbf{S}$. The state of the crowd, which consists of individual pedestrians, is a union of the set of each pedestrian's state $\mathbf{X} = \bigcup_i \mathbf{x_i}$, where subscript $i$ denotes the $i^{th}$ pedestrian. We do not explicitly model or capture pairwise interactions between pedestrians. However, the difference between $\mathbf{v}^{pref}$ and $\mathbf{v}^c$ provides partial information about the local interactions between a pedestrian and the rest of the environment.

**Motion Model:** $\mathbf{M} \in \mathbb{R}^5$ denotes the set of parameters of the motion model. The motion model corresponds to the local navigation rule or scheme that each pedestrian uses to avoid collisions with other pedestrians or obstacles. Our formulation is based on the RVO velocity-based motion model [37]. Our approach can also be combined with other models based on social forces or Boids. In this model, the motion of each pedestrian is governed by these five characteristics: *Neighbor Dist, Maximum Neighbors, Planning Horizon, Radius, and Preferred Speed*. We propose an algorithm to compute these motion model characteristics for each pedestrian.

### B. State Estimation

It is difficult to extract perfect trajectories from a real-world scenario. The extracted trajectories are likely to have incomplete tracks and noise [9]. To compensate for any errors, we use the Bayesian inference technique to compute the state of each pedestrian.

We assume that each pedestrian's true state is denoted by $\hat{\mathbf{x}}^\mathbf{t}$ and a tracking algorithm computes an observation of the pedestrian's position $\mathbf{z}^t \in \mathbb{R}^2$ at each timestep using an observation function $h()$ with sensor error $\mathbf{r} \in \mathbb{R}^2$. The sensor error is assumed to follow a zero-mean Gaussian distribution with covariance $\Sigma_r$:

$$z^t = h(\hat{\mathbf{x}}^t) + \mathbf{r}, \mathbf{r} \sim N(0, \Sigma_r). \quad (2)$$

Any tracking or synthetic algorithms that provide the trajectory of each pedestrian can be used as the observation function $h()$.

The true real-world crowd dynamics can be approximated by a state-transition model $f()$. The prediction error in $f()$ is represented as a zero-mean Gaussian distribution $\mathbf{q} \in \mathbb{R}^6$ with covariance $\Sigma_q$:

$$\mathbf{x}^{t+1} = f(\mathbf{x}^t) + \mathbf{q}, \ \mathbf{q} \sim N(0, \Sigma_q). \quad (3)$$

We estimate the most likely state $\mathbf{x}$ of each pedestrian using an Ensemble Kalman Filter (EnKF) and Expectation Maximization (EM) with the observation model $h()$ and the state transition model $f()$. EnKF uses a predictive scheme to provide a state estimate for a non-linear state-transition model, which corresponds to various components of a pedestrian's motion and behavior. During the prediction step, EnKF predicts the next state based on the underlying transition model and $\Sigma_q$. When a new observation based on real-world sensor data or the extracted trajectory parameters is available, $\Sigma_q$ is updated based on the difference between the observation and the prediction.

**EM for state estimation:** Expectation Maximization (EM) is an iterative process that maximizes the likelihood of the latent variable [23]. EM is an iterative process that repeats the E step, which computes the expected value for $\Sigma_q$ (in our case by using EnKF) and the M step, which computes the distribution with the computed value $\Sigma_q$ during the previous

E step. Ultimately, we use it to compute the best estimate of the state, for the given observation data. The underlying formulation is expressed as:

$$E(ll(\Sigma_q)) = -\sum_{t=0}^{t-1} E((\mathbf{x}^{t+1} - f(\mathbf{x}^t)^\mathbf{T} \Sigma_q^{-1}(\mathbf{x}^{t+1} - f(\mathbf{x}^t))).$$

We can estimate this value by finding the average error for each sample in the ensemble at each timestep for each agent. This pedestrian state information is used to classify the pedestrian behavior as well as compute the global features. In particular, we estimate the five RVO-based motion model parameters of each pedestrian. Note that these parameters can vary dynamically, as the pedestrian interacts with the obstacles and other pedestrians in the environment. Next, we use these parameters to estimate the personality traits.

## C. Personality Traits and Psychological Cues

A key issue in our approach is classifying the personality of each pedestrian in the crowd. Psychologists have proposed various ways of characterizing the personalities exhibited by pedestrians. These include modeling the variations in behavior using personality models based on cognitive models [13], OCEAN personality factors [8], the MBTI personality model [32], Personality Trait Theory [14], etc. Our formulation of psychological constraints is based on Trait Theories of Personalities, a theory that categorizes people's behavior based on a small number of personality traits [28]. We use this approach to automatically classify the personality or behavior of every pedestrian in a crowd. In particular, we characterize each pedestrian's behavior based on a weighted combination of different personality traits that are inferred from his/her movement pattern and interactions with other pedestrians and obstacles in the environment. We use the well-known Personality Trait Theory from psychology and the Eysenck 3-factor model [10] to classify such behaviors. This model identifies three major factors that characterize a personality: *Psychoticism*, *Extraversion*, and *Neuroticism* (commonly referred to as **PEN**). Each of these three traits has been linked to a biological basis, such as levels of testosterone, serotonin, and dopamine present in one's body, and we use them to characterize the psychological constraints.

In our case, each pedestrian's personality is identified based on how they exhibit each of these three traits and are classified into six weighted behavior classes: *aggressive, assertive, shy, active, tense,* and *impulsive*. We chose these six particular behavior characteristics because they are useful in describing the behaviors of pedestrians and span the space covered by the PEN model, with at least two adjectives for each PEN trait [28]. We classify each pedestrian's behavior using these traits because our fundamental assumption is that each pedestrian's behavior can be captured by a weighted combination of these six traits.

A key issue in this formulation is defining a mapping between the five motion model parameters for the RVO model described in Section III-B with these six personality traits. We make use of the data-driven mapping presented in [14] to derive a linear mapping that adopts the results of a user study as:

$$\mathbf{B} = \begin{pmatrix} w_1 * Aggressive \\ w_2 * Assertive \\ w_3 * Shy \\ w_4 * Active \\ w_5 * Tense \\ w_6 * Impulsive \end{pmatrix} = \mathbf{RVO}_{mat} * \begin{pmatrix} \frac{1}{13.5}(Neighbor\ Dist - 15) \\ \frac{1}{49.5}(Max.\ Neighbors - 10) \\ \frac{1}{14.5}(Planning\ Horiz. - 30) \\ \frac{1}{0.85}(Radius - 0.8) \\ \frac{1}{0.5}(Pref.\ Speed - 1.4) \end{pmatrix}$$

where,

$$\mathbf{RVO}_{\mathbf{mat}} = \begin{pmatrix} -0.02 & 0.32 & 0.13 & -0.41 & 1.02 \\ 0.03 & 0.22 & 0.11 & -0.28 & 1.05 \\ -0.04 & -0.08 & 0.02 & 0.58 & -0.88 \\ -0.06 & 0.04 & 0.04 & -0.16 & 1.07 \\ 0.10 & 0.07 & -0.08 & 0.19 & 0.15 \\ 0.03 & -0.15 & 0.03 & -0.23 & 0.23 \end{pmatrix}$$

and $w_i$ are personality weights for each of the six personalities. This mapping was derived based on user studies and perception of different behaviors that were generated by varying the motion parameters and performing statistical analysis on the user-generated data. We obtain the motion model parameters from the EM algorithm and then use the above mapping to determine the values of the personality traits, $\mathbf{b} = \{Aggressive, Assertive, Shy, Active, Tense, Impulsive\}$. We use the default $\mathbf{w}$ values ($w_i = 1$) for this step. Even though our approach is general, the mapping ($\mathbf{RVO_{mat}}$) is specific to the RVO motion model and the user study described in [14]. However, we can use different mappings or other forms of regression to compute such a mapping between the personality characteristics and the motion model. Next, we use these personality characteristics to compute proxemic distances for socially-aware navigation.

## D. Pedestrian Path Prediction

The key aspect of any real-time prediction algorithm is estimating the motion parameters of a pedestrian as they provide the best estimator of their movement in a dense setting. We represent the motion parameters of a pedestrian $i$ by his/her state $\mathbf{x_i}$. Given the state of the crowd $\mathbf{X} = \bigcup_i \mathbf{x_i}$ for the previous $n$ frames at time $t$, our pedestrian path prediction algorithm predicts the motion parameters of a pedestrian $i$ for a future time, $\mathbf{x_i^{t+\Delta t}}$. We compute these motion parameters from the personality trait learning module described above in Section III-C. We use these estimated motion parameters to extend the accuracy of the real-time prediction algorithm GLMP [4].

Pedestrian behavior may have slight variations during the course of the motion. To capture these variations, we find an upper bound $\mathbf{M}_{ub}$ and a lower bound $\mathbf{M}_{lb}$ on the motion parameters. From the computed values of six personality traits ($\mathbf{b}$) in Section III-C, we compute the trait with the largest value, the most dominant trait, $b_d$. $\mathbf{M}_{ub}$ is calculated by adding $y\%$ to $b_d$ and adding $(y/3)\%$ to the other traits ($\mathbf{b} \setminus b_d$). Similarly, we subtract $y\%$ and $(y/3)\%$ to compute $\mathbf{M}_{lb}$, where $y$ is a user-defined variable. A value of $y = 5\%$ is able to capture noise and the natural variance in the pedestrian behavior.

If the individual pedestrian parameters are within the bounds $\mathbf{M}_{lb}$ and $\mathbf{M}_{ub}$, we use them directly for prediction; otherwise, we clamp $\mathbf{M}$ to the corresponding boundary value of the motion parameters, ($\mathbf{M}_{lb}$, $\mathbf{M}_{ub}$). Having updated the motion model parameters, we recompute the personality weights $\mathbf{w}$. Our formulation assumes that each pedestrian behavior generally remains constant and lies within a range of variance; if the change in pedestrian behavior is substantial between successive frames, there is a possibility of error in prediction. Since the behaviour of a pedestrian might changes considerably over a long period of time, we re-sample the behavior every few frames.

$$\mathbf{M} = \begin{cases} \mathbf{M}_{lb}, & \text{if } \mathbf{M_i} \leq \mathbf{M_{lb_i}}, \forall i \\ \mathbf{M_{ub}}, & \text{if } \mathbf{M_i} \geq \mathbf{M_{ub_i}}, \forall i \\ \mathbf{M}, & \text{otherwise} \end{cases}$$

Next, we combine these motion parameters $\mathbf{M}$ with the local movement patterns from to perform long-term path prediction for each pedestrian.

## IV. SocioSense: Robot Navigation

### A. Proxemic Distances (Social Cues)

A key component of socially-aware navigation is computation of the appropriate distance between the robot and each pedestrian, based on the notion of proxemics [15]. In this context, we use the notions of *public distance* and *social distance* to perform socially-aware navigation. In particular, public distance refers to the distance at which people can give a speech and social distance characterizes the distance at which people can talk to each other. These distances have been known to vary according to cultural norms, environment, or an individual's personality. In our formulation, we mainly focus on variations in these distances originating from the differences in personality. It has been shown that traits like *Extraversion* affect interpersonal distances. Though social and public distances do not vary significantly, extroverts can have a smaller personal distance than introverts [39]. Based on the formulation described in [39], we compute the limits on an individual's personal distance (Table I). We obtain the personal distance of an individual by taking a weighted average of these limits, weighted by a computed value of *Extraversion* ($PEN_e$) for that individual.

|  | Personal Distance | Social Distance |
|---|---|---|
| *Extrovert* | 179.58 | 267.97 |
| *Introvert* | 88.9 | 233.17 |

**TABLE I:** *Extraversion vs Personal/Social Distances:* The personal distance indicates the minimum distance before the pedestrian feels uncomfortable with the robot. All distances are given in cms.

In order to determine the *Extraversion* of each pedestrian, we compute the mapping of the six personality traits (**b**, computed earlier) to a 3-factor **PEN** model based on the function described below:

$$\begin{pmatrix} \mathbf{P}sychoticism \\ \mathbf{E}xtraversion \\ \mathbf{N}euroticism \end{pmatrix} = \mathbf{PEN_{mat}} * \mathbf{B}$$

where,
$$\mathbf{PEN_{mat}} = \begin{pmatrix} 0.22 & 0.28 & -0.09 & -0.01 & -0.17 & 0.31 \\ 0.16 & 0.33 & 0.07 & 0.53 & 0.05 & 0.10 \\ -0.15 & 0.16 & 0.47 & -0.01 & 0.42 & -0.08 \end{pmatrix}.$$

Currently, we use a single scalar quantity, *Extraversion* ($PEN_e$), from the PEN model. Based on the proposed distances given in Table I, we establish a linear mapping between the pedestrians' personality traits and their personal distance to the robot. After normalizing $PEN_e$ by dividing it by the magnitude of the **PEN** vector, we compute the personal distance $d_p$ of the pedestrian:

$$d_p = 179.58(PEN_e) + 88.9(1 - PEN_e)$$

For the other distance metrics,, like social distance $d_s$, we select a distance from Table I. If $PEN_e < 0.5$ then $d_s = 233.17$ else $d_s = 267.97$.

### B. Socially-Aware Robot Navigation

We use the distances, $d_s$ and $d_p$, computed using the psychological constraints to enable socially-aware collision-free robot navigation through a crowd of pedestrians. Our navigation method is based on Generalized Velocity Obstacles (GVO) [38], which uses a combination of local and global methods. The global metric is based on a roadmap of the environment. The local method computes a new velocity for the robot and takes these distances into account. Moreover, we also take into account the dynamic constraints of the robot in this formulation.

Figure 4 illustrates how a robot avoids an approaching pedestrian based on these distances. At a given time instant, the pedestrian is located at $\mathbf{p}_{human}^{curr}$, and has two proxemic distances: a personal distance of $d_p$ (red) and a social distance of $d_p$ (yellow). At the same time instance, the robot is located at $\mathbf{p}_{robot}^{curr}$ and has a preferred velocity $\mathbf{v}_{robot}^{pref}$ that is computed based on global navigation module. This is the velocity that it would have for navigating to its goal position, in the absence of any static or dynamic obstacles. The robot predicts that during the next time frame, the pedestrian will move to the position $\mathbf{p}_{human}^{pred}$ using the path prediction described in Section III(D), and computes its new velocity to avoid a collision with the pedestrian. However this path prediction is not sufficient for socially-aware navigation, since the robot fails to take into account the pedestrian's proxemic distances. Based on these distances, the robot alters its goal position to $\mathbf{p}_{robot}^{pred+soc}$ and its velocity to $\mathbf{v}_{robot}^{pred+soc}$ to accommodate both social and psychological constraints. Notice that the velocity $\mathbf{v}_{robot}^{pred}$ causes the robot to intrude on the pedestrian's personal distance, shown by the red circle centered around the pedestrian, whereas the updated velocity $\mathbf{p}_{robot}^{pred+soc}$ successfully accounts for the pedestrian's personal distance and as well as its social distance.

We assume that the pedestrians in the environment are non-cooperative and may not actively avoid collisions in a reciprocal manner with the robot. Thus, the robot must assume 100% responsibility to avoid collisions and keep safe distances. During the navigation, our *SocioSense* algorithm predicts the pedestrian's future motion and avoids any steering inputs that result in a collision with the predicted pedestrian positions (Figure 3). Our approach is agnostic to the underlying navigation algorithm (e.g. GVO) and can be combined with other methods like potential field methods.

## V. Performance and Analysis

We have evaluated the accuracy of our real-time prediction algorithm (described in Section III-B) against other state of the art real-time methods. We applied our algorithm to the 2D pedestrian trajectories tracked from different video datasets, and calculated the accuracy of our predicted positions relative to the ground truth using an error metric. Because of error accumulation over time, trajectory prediction algorithms tend to be more accurate over shorter time windows as compared to longer time windows. Therefore, we measure the accuracy results for two time windows for each algorithm - short (1 second) and long (5 seconds). In terms of video benchmarks, we chose different crowd video datasets corresponding to indoor and outdoor scenes with varying pedestrian density and cultural background. The density varied considerably: from low (less than 1 pedestrians/m$^2$) to medium (1-2 pedestrians/m$^2$) to high (more than 2 pedestrians/m$^2$). Table II summarizes the crowd video datasets, their crowd characteristics, and the accuracies of the predicted trajectories computed by different real-time algorithms for both time windows. The last columns highlight the improved prediction results computed using our approach.

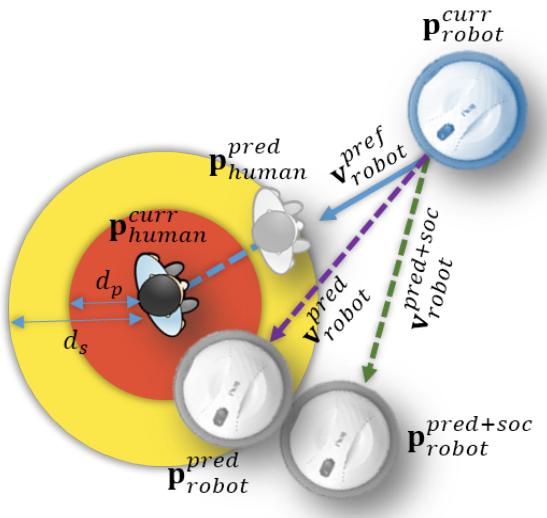

**Fig. 4:** *Our robot navigation algorithm satisfies the proxemic distance constraints, including personal space (red) and social space (yellow). The trajetory computed by our SocioSense navigation algorithm (green trajectory) does not intrude on the personal/social space of the pedestrian, whereas a robot that fails to take into account the social constraints (purple trajectory) may cause discomfort to the pedestrians.*

### A. Prediction Accuracy

We use a simple metric to evaluate the accuracy of our predicted trajectories. The average human stride length is about $0.8$ meters [31]. For a given time instant, the predicted pedestrian position is counted as successful when the estimated mean error between the predicted position and the ground truth value is less than this constant. We define prediction accuracy at a time instant as the ratio of the number of "successful" predictions and the total number of tracked pedestrians in the video.

A prediction algorithm should be able to predict trajectories over a long time horizon. All the algorithms listed in Table II have reduced accuracy scores when a longer time window is used. Even in these situation, our approach outperforms (or does as well as) the other methods in these crowd video datasets. In particular, our algorithm is significantly more accurate than competing real-time methods in predicting trajectories over a *long time window* in *high-density* crowd datasets like *Marathon* and *IITF-5*.

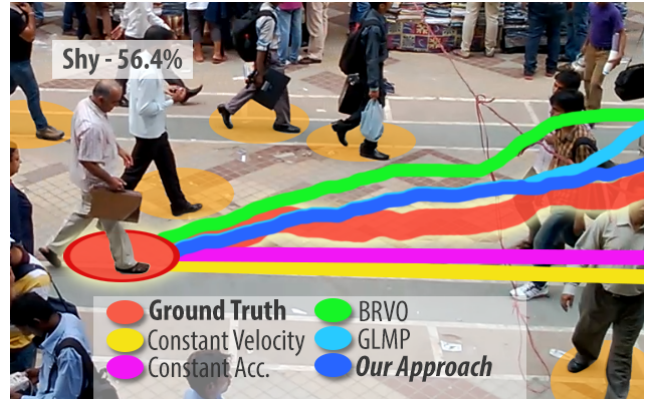

**Fig. 5:** *Our approach automatically classifies pedestrian behaviors in real-time (e.g. Shy behavior of a pedestrian). This behavior information is used to dynamically compute motion parameters and improve the performance of our long-term prediction algorithm (shown in dark blue) and compute proxemic distances. Our results are very close to ground truth (shown in red) and offer up to 21% improvement over prior real-time algorithms, whose predicted trajectories are shown in different colors. This demonstrates the benefits of using the psychological constraints for prediction and navigation.*

### B. Socially-aware Navigation

We evaluate the performance of our socially-aware navigation algorithm, *SocioSense* with other algorithms without that do not take into account proxemic or social constraints. We compute the number of times the non-social robot intrudes on the personal space of the pedestrians, and thereby results in discomfort for some of the pedestrians. We also measure the additional time a robot with our *SocioSense* algorithm takes to reach the goal position, without any intrusions of pedestrians' personal/social spaces. Our results (Table III) demonstrate that in $< 30\%$ additional time *SocioSense* can reach its goal while ensuring that the personal/social space of any pedestrian is not intruded. Table III also lists the time taken to compute proxemic social constraints.

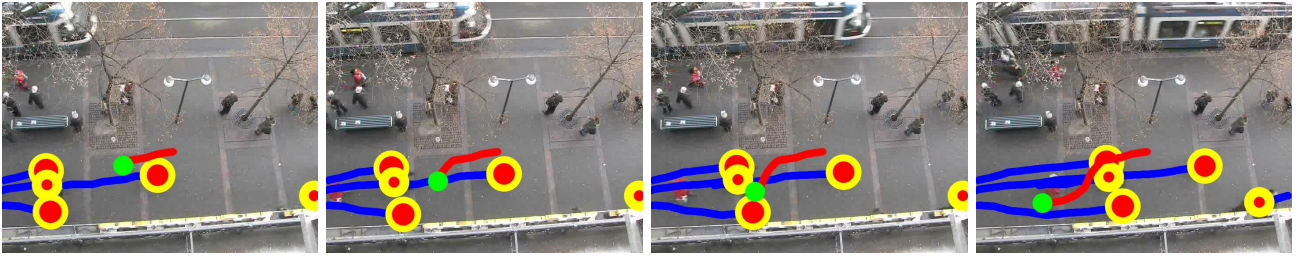

**Fig. 3:** *Example robot trajectory* navigating through the crowd in **Hotel** dataset. Red/yellow circles represent current pedestrian positions(personal/social distance), green circles are the current position of the robot.

| Dataset | Challenges | Density | # Tracked | ConstVelocity | | Kalman Filter | | GLMP (Baseline) | | SocioSense | |
|---|---|---|---|---|---|---|---|---|---|---|---|
| | | | | *1 sec* | *5 sec* | *1 sec* | *5 sec* | *1 sec* | *5 sec* | *1 sec* | *5 sec* |
| UCSD-Peds1 | BV, PO, IC | Low | 43 | 75.2% | 54.1% | 76.4% | 56.4% | 79.1% | 61.6% | 84.3% | 74.3% |
| Marathon | BV, PO, IC, CO | High | 76 | 32.1% | 10.4% | 35.2% | 11.3% | 41.3% | 17.2% | 46.8% | 21.1% |
| NDLS-2 | BV, PO, IC | High | 121 | 57.8% | 31.0% | 52.0% | 34.1% | 59.7% | 39.6% | 68.2% | 43.0% |
| MANKO | BV, PO, IC, CO | High | 87 | 43.7% | 19.4% | 44.4% | 19.7% | 49.4% | 24.0% | 59.2% | 28.9% |
| 879-38 | BV, PO, IC, CO | High | 43 | 38.1% | 29.1% | 38.7% | 31.0% | 41.2% | 35.6% | 48.1% | 42.9% |
| Crossing | BV, PO, IC | High | 37 | 75.3% | 52.0% | 76.1% | 53.1% | 79.0% | 59.3% | 86.6% | 59.3% |
| IITF-1 | BV, PO, IC, CO | High | 167 | 63.5% | 33.4% | 63.9% | 39.1% | 65.3% | 41.8% | 65.3% | 41.8% |
| IITF-3 | BV, PO, IC, CO | High | 189 | 61.1% | 29.1% | 63.6% | 31.0% | 67.6% | 37.5% | 78.2% | 42.9% |
| IITF-5 | BV, PO, IC, CO | High | 71 | 59.2% | 28.8% | 61.7% | 29.1% | 62.9% | 30.1% | 68.1% | 34.0% |
| NPLC-1 | BV, PO, IC | Medium | 79 | 76.1% | 63.9% | 78.2% | 65.8% | 79.9% | 69.0% | 81.2% | 72.6% |
| NPLC-3 | BV, PO, IC, CO | Medium | 144 | 77.9% | 70.1% | 79.1% | 71.9% | 80.8% | 74.4% | 88.1% | 74.6% |
| Students | BV, IC, PO | Medium | 65 | 65.0% | 58.2% | 66.9% | 61.0% | 69.1% | 63.6% | 69.1% | 63.6% |
| Campus | BV, IC, PO | Medium | 78 | 62.4% | 57.1% | 63.5% | 59.0% | 66.4% | 59.1% | 66.4% | 59.1% |
| seq_hotel | IC, PO | Low | 390 | 74.7% | 67.8% | 76.7% | 68.3% | 76.9% | 69.2% | 81.2% | 73.6% |
| Street | IC, PO | Low | 34 | 78.1% | 70.9% | 78.9% | 71.0% | 81.4% | 71.2% | 81.4% | 71.2% |

**TABLE II:** *Accuracy Benchmarks:* We compare our path prediction algorithm with state of the art real-time algorithms, on crowd video datasets with varying densities and numbers of tracked pedestrians, and time windows of 1 sec and 5 sec. Our approach, SocioSense, consistently outperforms the other methods, even for challenging datasets like Marathon. Abbreviations used for scene characteristics: BV: Background Variations, PO: Partial Occlusion, CO: Complete Occlusion, IC: Illumination Changes.

| Dataset | Additional Time | Performance | Intrusions Avoided |
|---|---|---|---|
| UCSD-Peds1 | 27% | 3.00E-04 ms | 11 |
| NDLS-2 | 13% | 2.74E-04 ms | 24 |
| Students | 11% | 0.72E-04 ms | 17 |
| seq_hotel | 17% | 0.98E-04 ms | 31 |

**TABLE III:** *Navigation Performance:* A robot using our SocioSense navigation algorithm can reach its goal position, while ensuring that the personal/social space of any pedestrian is not intruded with $< 30\%$ overhead. We evaluated this performance in a simulated environment, though the pedestrian trajectories were extracted from the original video.

## VI. LIMITATIONS, CONCLUSIONS, AND FUTURE WORK

We present a novel algorithm *SocioSense* that performs socially-aware navigation based on psychological as well as social constraints. We use Bayesian inference along with Personality Trait theory to identify the time-varying personality of each pedestrian in the video. These behaviors are used for path prediction as well as computing proxemic distances for social navigation. . We highlight the performance benefits of our prediction algorithms on complex benchmark datasets, and compare their accuracy against state of the art real-time prediction algorithms. Additionally, we demonstrate that our *SocioSense* algorithm avoids intrusions on the personal/social space of pedestrians.

Our approach has some limitations. The behavior classification is based on personality models and the Eysenck PEN model, which may not be sufficient to capture all observed behaviors. One possibility is to combine the classification scheme with other personality models like the Myers-Briggs Type Indicator [24]. Furthermore, a behavior representation based on six traits may not be sufficient. We would like to integrate these algorithms with different robots and evaluate their performance in crowded indoor and outdoor scenes. We would also like to take into cultural norms and group behaviors during social-aware navigation.